# Cultural Adaptation in Large Language Models for Political Discourse

**Wajdi Zaghouani**
Northwestern University in Qatar
wajdi.zaghouani@northwestern.edu

**Abstract**

The integration of large language models into political discourse analysis creates new opportunities for comparative research, policy analysis, and civic technology, while introducing material risks for democratic accountability. This paper argues that cultural adaptation is a prerequisite for trustworthy deployment of large language models in political communication across diverse linguistic and institutional contexts. Current systems remain shaped by English dominant data, uneven multilingual coverage, and assumptions grounded in a narrow range of political institutions and discourse conventions, producing systematic errors when applied across cultures. We formalize cultural adaptation across translation, discourse, and ontology levels, identify recurring cultural failure modes in political NLP, and propose an operational evaluation matrix grounded in cultural fidelity, calibration, and democratic safety. Building on political text analysis, sociotechnical auditing, and cross cultural pragmatics, we outline methodological pathways including participatory dataset development, culturally aware transfer learning, and benchmark design that makes cultural adaptation empirically measurable. We conclude by clarifying governance constraints and scope conditions under which culturally adaptive political NLP can support democratic legitimacy.

## 1. Introduction

Political communication is increasingly mediated by generative language technologies. Large language models are now used for moderation support, narrative analysis, translation and summarization of public input, and exploratory policy analysis. These uses place language models inside workflows that shape what is visible, what is amplified, and what is treated as credible. As a result, model limitations are not only technical errors. They can become representational failures with direct implications for legitimacy and inclusion.

A central tension is that large language models are often presented as broadly general, yet they are trained on data that is uneven across languages, regions, and political traditions. Work on dataset documentation and model reporting shows that claims of generality should be supported by explicit statements about data provenance, intended use, and known failure modes (Bender and Friedman, 2018; Mitchell et al., 2019; Gebru et al., 2021). Critical analyses of scaling trends also emphasize that larger models can amplify harms when deployed without clear governance and evaluation in the settings where they will matter (Bender et al., 2021).

Political language is a particularly demanding domain because it is strategic, contested, and context dependent. The same surface form can function as mobilization, satire, coded dissent, or institutional signaling depending on local conventions and power relations. Decades of research in political text analysis show both the promise and the pitfalls of treating text as data. Early warnings about measurement validity and domain shift remain relevant in the era of large language models (Grimmer and Stewart, 2013; Gentzkow et al., 2019).

This paper focuses on culturally adaptive large language models for political discourse and cross cultural reasoning. We define cultural adaptation as more than translation. It includes pragmatic competence, genre awareness, and the ability to represent political concepts as they are used within a local political culture (Wierzbicka, 1991; Brown and Levinson, 1987). We contribute three elements. First, we describe recurring cultural failure modes when large language models are used for political discourse analysis. Second, we propose a framework for culturally aware development and evaluation that integrates pragmatics, political theory, and sociotechnical governance. Third, we outline transparency and accountability mechanisms aligned with research on auditing, documentation, and explainability (Ribeiro et al., 2016; Raji et al., 2020).

## 2. Background and Related Work

### 2.1. Political text as data and domain specific validity

Political scientists have long used textual data to infer preferences, ideologies, and issue attention. Classical scaling approaches such as Wordscores demonstrate how word usage can be mapped to latent political dimensions, while underscoring the dependence of inference on reference texts and corpus construction (Laver et al., 2003; Lowe, 2008).

Applications to legislative speech and party documents further show that institutional setting and genre strongly condition model behavior (Slapin and Proksch, 2008).

The broader shift toward treating text as data introduced methodological clarity around preprocessing, measurement error, and validation. Research in Political Analysis emphasizes that held out accuracy is insufficient for substantive inference, particularly under domain shift and strategic adaptation by political actors (Grimmer and Stewart, 2013; Denny and Spirling, 2018). These concerns extend to large language models, where fluency can obscure brittleness, poor calibration, and systematic omissions in underrepresented contexts.

### 2.2. Cross cultural pragmatics and political discourse conventions

Political meaning depends on norms governing indirectness, evidentiality, authority, and persuasive form. Cross cultural pragmatics documents variation in speech acts, implicatures, and politeness strategies across communities (Wierzbicka, 1991; Brown and Levinson, 1987). Because many political NLP tasks such as stance detection, toxicity classification, and misinformation triage rely on pragmatic inference rather than literal semantics, these differences are consequential.

Institutional and historical factors amplify pragmatic variation. Some political cultures privilege adversarial argumentation, while others emphasize narrative testimony, moral evaluation, or communal authority. Models trained on one dominant discourse tradition may systematically misinterpret others. Such misreadings are not neutral errors but can marginalize culturally legitimate forms of political reasoning.

### 2.3. Large language models, cultural representation, and linguistic diversity

Transformer architectures and large scale pretraining substantially improved NLP performance (Vaswani et al., 2017; Devlin et al., 2019). Multilingual models extended this paradigm across languages, yet evaluations reveal uneven cross lingual transfer, with substantial variation by language family and resource level (Pires et al., 2019; Hu et al., 2020). Analyses of linguistic diversity further show that incentives and benchmarks concentrate attention on a limited subset of languages despite multilingual claims (Joshi et al., 2020).

Representation concerns extend beyond language coverage. WEIRD sampling critiques illustrate how data pipelines can encode narrow demographic perspectives while appearing universal (Henrich et al., 2010). Surveys of bias in NLP show how training data, annotation practices, and evaluation design shape model behavior (Blodgett et al., 2020; Hovy and Spruit, 2016). In political contexts, such gaps can distort analyses of ideology and issue salience, producing concept mappings that misalign across political systems. These failures are ontological as well as lexical.

### 2.4. Documentation, explainability, and algorithmic accountability

Trustworthy political NLP requires transparency regarding data provenance, intended use, and limitations. Documentation frameworks including data statements, datasheets, and model cards operationalize this transparency (Bender and Friedman, 2018; Mitchell et al., 2019; Gebru et al., 2021). Broader analyses of dataset construction show how documentation failures propagate harms across the pipeline (Paullada et al., 2021).

Explainability methods such as local surrogate models and feature attribution support interpretability (Ribeiro et al., 2016; Lundberg and Lee, 2017). However, interpretability alone does not ensure accountability. Auditing frameworks emphasize organizational processes for monitoring, contestation, and remediation after deployment (Raji et al., 2020).

## 3. A Framework for Cultural Adaptability in Political NLP

### 3.1. Levels of cultural engagement

We distinguish three levels of cultural engagement in political language technology. Translation level adaptation renders political content into another language while preserving propositional meaning. Discourse level adaptation accounts for local genre conventions, rhetorical strategies, and pragmatic norms. Ontology level adaptation represents political concepts, institutions, and normative categories in culturally grounded ways, including when concepts do not align cleanly across contexts.

Large language models commonly excel at translation level adaptation and can sometimes mimic discourse level patterns. Ontology level adaptation is far less reliable because it depends on concept mappings that are rarely explicit in training data. Ontology level adaptation is also the level most directly tied to legitimacy concerns, because it governs what the model treats as political evidence, what it counts as justification, and what it implicitly normalizes as a proper political order.

### 3.2. Cultural failure modes and diagnostic cases

Cultural failure modes in political NLP often appear as ordinary classification errors, but their structure is systematic. One common pattern is pragmatic misreading, where indirect speech, irony, satire, or honorific address is treated as literal endorsement. Another pattern is genre mismatch, where rhetorical forms such as sermons, oral testimony, or poetic political expression are mapped to incorrect categories because the model expects the structure of parliamentary debate or journalistic prose.

A third pattern is concept collapse. Political concepts that appear similar across contexts are treated as equivalent even when they correspond to different institutional arrangements, moral vocabularies, or historical trajectories. Concept collapse is especially likely when the model relies on dominant language sources, because it will anchor meanings in those sources. The risk parallels what Sartori (1970) diagnosed as "concept stretching" in comparative politics: when categories developed in one institutional setting are applied uncritically to others, they lose discriminatory power. For example, Schaffer's fieldwork on Wolof speakers in Senegal showed that the local concept of *demokaraasi* foregrounds collective security and community solidarity rather than individual choice among competing candidates, a meaning that diverges substantially from the liberal procedural conceptions encoded in most English language training data (Schaffer, 1998). Similarly, the Arabic term *shura* (consultation) carries religious and communal authority dimensions that are absent from "deliberation" as used in Western democratic theory. A stance detection model trained on English parliamentary data that encounters *shura* based reasoning in Arabic political discourse is likely to misclassify the pragmatic function of such arguments. Work on multi dialectal Arabic stance detection across political contexts confirms that annotation categories calibrated for one political culture do not transfer reliably to others, even within a single language family (Charfi et al., 2024b).

These failure modes suggest that evaluation should include targeted diagnostic suites. Multilingual benchmarks such as XTREME provide broad coverage for cross lingual generalization, but they are designed for general purpose NLP tasks such as classification, structured prediction, and question answering, and they do not probe culturally specific political ontologies, pragmatic conventions, or genre variation (Hu et al., 2020). Political NLP also requires domain specific suites that test pragmatic competence, genre recognition, and concept mappings.

### 3.3. Participatory dataset development and cultural governance

Cultural adaptation requires data that reflects local genres, political institutions, and community defined categories. This is not only a sampling problem. It is a governance problem because political texts can be sensitive and because annotation choices encode political assumptions. Datasheets and data statements can support transparency about what is included and what is missing, and can help surface mismatches between a dataset and the claims made from it (Bender and Friedman, 2018; Gebru et al., 2021).

Participatory dataset development goes further by involving relevant communities in selecting genres, defining labels, and validating edge cases. Participatory approaches help reduce misrepresentation by recognizing that political meaning is negotiated and that local interpretive expertise cannot be fully captured by generic label schemes. This perspective aligns with data governance principles that emphasize community rights and responsibilities over how data is collected and used, especially for historically marginalized groups (Carroll et al., 2020).

Participatory governance can be operationalized through documentation, review boards that include community representatives, clear consent and access rules, and mechanisms for redress. These mechanisms are particularly important when data involves vulnerable speakers or politically risky contexts.

### 3.4. Culturally aware transfer learning

Cross cultural transfer should be selective rather than assumed. Some linguistic knowledge transfers broadly, such as syntactic regularities and general discourse structure. Political ontologies and culturally specific pragmatic cues often do not. Multilingual evaluation shows that cross lingual transfer can vary widely across languages even on general tasks, suggesting caution when transferring political category systems (Pires et al., 2019; Hu et al., 2020).

We recommend transfer setups that separate general language competence from local political calibration. This motivates mixed training protocols that combine multilingual representations with smaller culturally grounded corpora, as well as adaptation methods that emphasize calibration and uncertainty estimation rather than only accuracy.

### 3.5. Design implications for culturally adaptive architectures

Architecture choices can support cultural adaptation when paired with governance and evaluation.

Retrieval augmented generation can improve factual grounding when retrieval corpora are local, curated, and documented. However, retrieval can also import local biases and exclusion if the corpus is narrow. Similarly, instruction tuning and preference optimization can align models to a target discourse style, but can also impose external norms if alignment data is not locally grounded.

The practical implication is that cultural adaptation should be treated as a system property that depends on data, modeling, and deployment context. Documentation should reflect these dependencies rather than presenting the model as a context free artifact (Mitchell et al., 2019; Gebru et al., 2021).

# 4. Trustworthy Political NLP: Bias, Misinformation, and Ethical Risk

## 4.1. Bias as a democratic risk

Bias in political NLP is not only a fairness issue for individuals. It can shape whose political speech is counted as legitimate, whose claims are treated as credible, and which communities are disproportionately flagged for moderation or suspicion. Bias can enter through training data composition, annotation guidelines grounded in one political culture, and evaluation designs that reward conformity to dominant discourse norms (Blodgett et al., 2020; Hovy and Spruit, 2016). Research on multi label hate speech annotation in Arabic illustrates how label definitions calibrated for one cultural context can produce systematic disagreements when applied to dialectally diverse data, underscoring the need for culturally sensitive annotation protocols (Zaghouani et al., 2024).

Because political language is strategic, models also face distribution shift driven by actors adapting to platform rules. This dynamic raises validity concerns familiar from earlier political text methods, where measurement can change behavior and thereby change the signal being measured (Grimmer and Stewart, 2013).

## 4.2. Stereotypes, representational harms, and embedding based bias

Studies of distributional semantics show that learned representations can encode human like stereotypes, affecting downstream decisions in ways that are difficult to detect through aggregate accuracy metrics (Caliskan et al., 2017). In political analysis, such biases can surface as systematic differences in how models describe groups, attribute agency, or characterize violence and legitimacy. These risks are amplified in generative settings, where model outputs can appear authoritative and can be reused at scale.

The stochastic parrots critique emphasizes that scaling can intensify these issues when training data is scraped without adequate attention to consent, representativeness, and downstream harms (Bender et al., 2021). For political applications, a key governance requirement is therefore a shift from implicit assumptions to explicit risk assessment and documentation.

## 4.3. Misinformation, propaganda, and context dependence

Misinformation detection is often framed as classification, yet political truth claims are contested, evolving, and deeply contextual. Research on the spread of false information online shows that diffusion dynamics can differ between true and false content, but content level detection remains challenging and sensitive to domain shift (Vosoughi et al., 2018; Shu et al., 2017). Cognitive studies further suggest that vulnerability to partisan misinformation can reflect limited analytic reasoning rather than simple ideological bias, complicating interventions based solely on content filtering (Pennycook and Rand, 2019).

Propaganda detection research illustrates both opportunities and limitations of automated approaches. Shared tasks on fine grained propaganda detection have produced datasets and modeling strategies, but their labels and examples are drawn from specific media and political contexts, and their transfer to other contexts is not guaranteed (Da San Martino et al., 2019).

Trustworthy political NLP therefore requires context aware pipelines that separate at least three activities: factual verification, rhetorical analysis, and normative judgment. Each activity has different evidentiary standards and different accountability requirements.

## 4.4. Explainability, uncertainty, and contestability

When political NLP systems are used in moderation, analysis, or policy workflows, users need to understand why a system produced an output and how reliable it is. Local explanation methods can increase transparency by showing which inputs most influenced a prediction (Ribeiro et al., 2016). Attribution methods such as SHAP can also support debugging and auditing, although they can be misinterpreted if used as proof of causal reasoning rather than as a diagnostic (Lundberg and Lee, 2017).

Transparency should be paired with contestability. Auditing frameworks propose end to end processes that include documentation, stakeholder review, monitoring, and pathways for appeal when outputs affect rights or participation (Raji et al., 2020). For political settings, contestability is not a purely technical feature. It requires institutional processes and clear authority to revise models, thresholds, and label definitions.

# 5. Generative AI for Deliberation and Policy Work

## 5.1. Deliberative democracy as a systems problem

Deliberative democratic theory grounds legitimacy in inclusive, reason giving public discussion, while deliberative systems theory extends this beyond single forums to the broader ecology of institutions and publics shaping collective reasoning (Mansbridge et al., 2012). Deliberation is thus distributed across media, civil society, legislatures, courts, and everyday discourse.

Large language models intersect with this system by summarizing inputs, translating across languages, clustering arguments, and assisting navigation of complex information. At the same time, they may distort deliberation by amplifying dominant frames, flattening disagreement, or generating persuasive text detached from accountable sources.

## 5.2. Augmentation rather than substitution

In deliberative contexts, generative systems should augment rather than replace human judgment. They can support scale through summarization, translation, and synthesis, but democratic legitimacy depends on accountable human decision processes. System design should therefore preserve agency, expose uncertainty, and ensure traceability from outputs to source segments.

Research on public consultation and deliberative polling shows that structured discussion improves opinion quality when participants receive balanced information and opportunities for reflection (Fishkin, 2009). Technologies that assist deliberation should thus be evaluated not only for linguistic performance but also for their effects on inclusion, balance, and mutual understanding.

## 5.3. Policy narratives and distributional blind spots

Generative models can map competing narratives, identify rhetorical strategies, and simulate policy justifications. However, training on historical corpora risks reproducing exclusions, omitting minority perspectives, and overgeneralizing dominant media frames. Because policy analysis informs resource allocation, systematic underrepresentation can produce material harms.

A practical safeguard is to treat generative outputs as hypotheses requiring validation through expert review, participatory feedback, and auditing, accompanied by documentation clarifying which publics and sources are represented in the underlying data (Mitchell et al., 2019; Gebru et al., 2021).

# 6. Governance, Regulation, and Institutional Accountability

## 6.1. Risk based governance for political applications

Political applications of AI can fall into higher impact categories because they may affect participation, access to information, and institutional decision making. Risk based governance approaches emphasize technical documentation, data governance, monitoring, and human oversight. Even when not legally mandated, these practices function as baseline requirements for trustworthy deployment in public facing settings.

Documentation frameworks provide the starting point by making system assumptions inspectable (Bender and Friedman, 2018; Mitchell et al., 2019). Auditing frameworks add an organizational layer that clarifies responsibility for monitoring, escalation, and remediation (Raji et al., 2020).

## 6.2. Transparency about training data and model limitations

Claims about cultural adaptability should be supported by evidence about training data coverage and by disaggregated evaluation results. Broad labels such as “multilingual” or “culturally aware” are not meaningful without clarity about the languages, dialects, genres, and political institutions represented. This is especially important because generative systems can produce plausible text even when their underlying concept mappings are incorrect.

The combination of data statements, datasheets, and model cards can document what the system was designed to do, where it is expected to fail, and what forms of oversight are necessary (Bender and Friedman, 2018; Mitchell et al., 2019; Gebru et al., 2021).

## 6.3. Environmental and labor considerations

Responsible political NLP should also document environmental and labor impacts. Research on the energy and carbon costs of deep learning in NLP highlights how training and experimentation decisions carry material externalities (Strubell et al., 2019). Political applications that claim social benefit should be explicit about these tradeoffs and should prioritize efficient adaptation methods where possible.

## 7. Evaluation, Reproducibility, and Responsible Practice

### 7.1. Beyond accuracy: cultural fidelity and democratic safety

Traditional metrics such as accuracy and F1 capture only narrow aspects of performance. For culturally adaptive political NLP, evaluation should include at least three families of measures. First, cross cultural robustness, which tests stability across languages, genres, and institutional contexts. Second, cultural fidelity, which assesses alignment with locally grounded meanings for political concepts and discourse acts. Third, democratic safety, which examines error patterns that affect participation, moderation, or informational access.

Multilingual benchmarks such as XTREME provide a model for disaggregated evaluation across languages and tasks (Hu et al., 2020). However, XTREME and similar resources evaluate general purpose NLP capabilities rather than culturally specific political reasoning. Their task designs do not distinguish between pragmatic competence and surface level accuracy, and they aggregate performance across languages without probing whether concept level meanings are preserved. These limitations make it difficult to detect the cultural failure modes described in Section 3.2, because a system can score well on general cross lingual transfer while still collapsing political ontologies across contexts. Political domain evaluation suites should therefore extend the disaggregated reporting model of XTREME while adding probes for pragmatic competence, genre recognition, and concept mappings.

### 7.2. Reproducibility with sensitive political data

Open science is more difficult when data is politically sensitive, but reproducibility remains essential for accountability and cumulative knowledge. The FAIR principles provide guidance for making research artifacts findable, accessible, interoperable, and reusable, with appropriate attention to constraints (Wilkinson et al., 2016). For communities with sovereignty concerns, governance principles such as CARE complement FAIR by emphasizing collective benefit, authority to control, responsibility, and ethics (Carroll et al., 2020).

Practical reproducibility strategies include tiered access, clear licensing, careful redaction, and well documented synthetic or diagnostic evaluation sets. When model weights or training data cannot be shared, thorough documentation and standardized reporting become more important, not less (Mitchell et al., 2019; Gebru et al., 2021).

### 7.3. Reporting standards for political NLP

Reporting should include corpus description, annotation procedures, inter annotator processes where relevant, model configuration, and disaggregated results by language and genre. It should also include qualitative error analysis focused on cultural failure modes. In political settings, qualitative analysis can reveal representational harms that aggregate metrics obscure.

### 7.4. Operational Evaluation Matrix for Cultural Adaptation

This subsection translates the paper's evaluation goals into an operational matrix that can be used to design benchmarks, audits, and acceptance criteria for political NLP systems deployed across cultural contexts. The matrix separates what is being evaluated, the unit of evaluation, the typical failure signal, and recommended measurement protocols. It also clarifies which evaluation artifacts support post deployment monitoring and contestation.

As summarized in Table 1, the proposed evaluation matrix specifies core dimensions, typical failure signals, and corresponding measurement protocols for culturally adaptive political NLP systems.

**Illustrative application to cross lingual stance detection.** To illustrate how the matrix functions in practice, consider a cross lingual stance detection system evaluated on political comments in German, French, and Italian, such as those represented in the X-Stance dataset (Vamvas and Sennrich, 2020). Under the *pragmatic competence* dimension, the evaluation would test whether the system correctly interprets indirect opposition expressed through rhetorical questions or hedged disagreement, which vary across Swiss language communities. Under *ontology alignment*, evaluators would check whether the system treats federalism related concepts consistently when the same political question is posed across languages, given that terms like “cantonal autonomy” carry different institutional weight in German speaking and French speaking Swiss political discourse. Under *calibration*, the evaluation would measure whether the system's confidence is appropriately lower for Italian inputs, where training data is sparser, compared to German inputs. Under *democratic safety*, auditors would examine whether stance detection errors disproportionately affect minority language speakers or speakers from underrepresented cantons. This example shows that the matrix does not replace standard accuracy evaluation but supplements it with structured diagnostics that make cultural adaptation claims falsifiable.

| Dimension | Evaluation unit | Failure signal | Methods and metrics |
|---|---|---|---|
| Pragmatic competence | Speech act and implicature | Indirect dissent interpreted as agreement, irony treated as literal stance, honorifics misread as endorsement | Curated pragmatic diagnostic set with local annotators. Report confusion patterns by speech act type. Add minimal pairs that differ only in pragmatic cue. |
| Genre recognition | Discourse genre and institutional setting | Sermon, oral testimony, or poetic political expression mapped to categories calibrated for parliamentary debate or journalistic prose | Cross genre evaluation suite. Report cross genre confusion matrix and error clustering by genre. Include genre shift stress tests with matched topics. |
| Ontology alignment | Political concepts and category mapping | Concept collapse across contexts, false equivalence of institutions, imported normative defaults | Expert adjudication protocol with concept mapping tasks. Report agreement on concept equivalence plus qualitative error taxonomy. Add retrieval grounded concept definitions when available. |
| Multilingual robustness | Language, dialect, and register | Performance cliffs for specific language varieties, systematic omissions, brittle transfer under code switching | Disaggregated metrics by language and dialect. Evaluate under translation, direct input, and code switching conditions. Include robustness tests with controlled lexical substitution. |
| Calibration and uncertainty | Confidence and abstention behavior | Overconfident errors in underrepresented contexts, failure to abstain under ambiguity | Calibration evaluation with expected calibration error and selective prediction curves. Measure abstention rates under ambiguity prompts and domain shift stress tests. |
| Cultural fidelity | Locally grounded meaning preservation | Summaries erase dissent, flatten moral vocabulary, or remove culturally salient argumentative forms | Human evaluation rubric with criteria for meaning preservation, dissent preservation, and culturally salient framing. Include citation based traceability from outputs to sources. |
| Democratic safety | Participation and moderation externalities | Disproportionate flagging, differential visibility, or systematic delegitimation of specific publics | Audit style evaluation with group conditioned error rates and disparity ratios. Include harm oriented incident taxonomy and red team scenarios relevant to local politics. |
| Contestability and accountability | Appeal and correction pathways | No actionable explanation, no route to challenge outputs, no remediation loop | Documentation and process checks. Require model cards and data statements. Track appeal outcomes, correction latency, and recurrence of known failure modes after remediation. |

Table 1: Operational evaluation matrix for culturally adaptive political NLP systems. The matrix specifies evaluation dimensions, units, common failure signals, and recommended methods. It complements accuracy oriented evaluation with diagnostics for cultural meaning, legitimacy, and governance.

**Weighting dimensions for deployment context.** The nine evaluation dimensions should not receive uniform weight across all deployment contexts. In civic technology applications such as public consultation summarization, cultural fidelity and democratic safety should take priority, because errors in meaning preservation or differential visibility directly affect civic participation. In academic research settings, ontology alignment and calibration may warrant greater emphasis, since the goal is to produce valid cross cultural comparisons. For commercial content moderation, pragmatic competence and multilingual robustness may be most critical, given the volume and linguistic diversity of inputs. We recommend that evaluation plans explicitly document the weighting rationale, and that any deployment above minimal risk include at least three dimensions evaluated with both quantitative and qualitative methods.

**Minimal reporting requirements.** At minimum, studies should report disaggregated results by language and genre, include qualitative error analysis tied to at least two culturally specific failure modes, and document the governance mechanisms that enable contestation and remediation. These requirements make claims of cultural adaptation falsifiable rather than aspirational.

## 8. Research Agenda and Methodological Pathways

We organize the research agenda into near term priorities that can be pursued with existing resources and longer term goals that require sustained community coordination.

### 8.1. Near term priorities

**Culturally grounded diagnostic benchmarks.** A first priority is the construction of diagnostic evaluation suites that probe cultural adaptation in political NLP. Several existing resources provide useful starting points. The X-Stance dataset covers more than 150 political questions in German, French, and Italian with 67,000 comments, enabling cross lingual and cross target stance evaluation (Vamvas and Sennrich, 2020). Multi dialectal Arabic resources such as MARASTA, which provides cross domain stance annotations across multiple Arabic dialects, demonstrate how annotation categories can be adapted to capture dialect specific pragmatic variation within a single language (Charfi et al., 2024b). Work on multi dialectal hate speech corpora further shows that culturally grounded annotation protocols are both feasible and necessary for reliable detection across dialect boundaries (Charfi et al., 2024a). Building on such resources, new diagnostic suites should incorporate pragmatic minimal pairs testing indirect dissent, irony, honorific address, and culturally specific politeness strategies; genre controlled splits comparing parliamentary debate, civic testimony, religious discourse, activist speech, and journalistic commentary within matched issue domains; and concept mapping tasks that require systems to distinguish partially overlapping political categories across contexts rather than collapsing them.

**Annotation protocols and ontology level documentation.** Benchmark construction should document how political categories are defined within each context, where categories are intentionally non equivalent across contexts, and how annotator disagreement is treated as epistemic signal rather than noise. Structured annotation guidelines may incorporate community review panels or domain experts to validate concept mappings, particularly in sensitive domains such as migration, minority rights, or sovereignty. Such documentation prevents implicit ontology transfer and makes alignment claims inspectable.

### 8.2. Longer term community goals

**Cultural ontologies and concept mapping.** A core research problem is how to represent political concepts in ways that preserve cultural specificity while supporting comparison. This includes mapping between concepts that are partially overlapping, detecting when translation hides key distinctions, and modeling how political terms shift meaning across institutions and time. Ontology level adaptation likely requires integrating linguistic representation learning with structured knowledge sources that are locally grounded and documented. The challenge is well illustrated by Sartori's ladder of abstraction: as political concepts travel across contexts, they must either gain generality at the cost of discriminatory power, or maintain specificity at the cost of comparability (Sartori, 1970). Computational approaches to this tradeoff remain largely unexplored.

**Participatory evaluation and community review.** Participatory evaluation can surface failures that are invisible to outside auditors. Community review processes can evaluate whether model outputs respect local meanings, avoid stereotyping, and provide useful summaries without erasing dissent. These processes also support legitimacy because they create channels for those affected to contest and correct system behavior.

**Shared tasks for democratic safety and reproducibility.** Beyond traditional classification, PoliticalNLP shared tasks can evaluate calibration under domain shift across languages and institutional settings, disparity in moderation or stance detection errors across demographic or linguistic groups, and preservation of dissent and minority viewpoints in summarization. Because political data is often sensitive, benchmark infrastructures should incorporate tiered access and transparent documentation aligned with FAIR and CARE principles (Wilkinson et al., 2016; Carroll et al., 2020). When full corpora cannot be released, diagnostic subsets, synthetic evaluation sets, and standardized reporting templates can enable reproducible comparison while respecting governance constraints.

**Auditing and continuous monitoring.** Political discourse shifts quickly in response to events, policies, and platform changes. Systems that are accurate at one time can drift into failure as vocabularies evolve and as actors adapt. Auditing frameworks emphasize continuous monitoring, incident reporting, and structured remediation processes (Raji et al., 2020). For political deployments, monitoring should include cultural drift indicators, not only statistical drift, because subtle shifts in pragmatics and framing can change interpretation.

## 9. Scope Conditions and Limits of Cultural Adaptation

Cultural adaptation is a necessary condition for trustworthy political NLP, but it is neither unlimited nor normatively neutral. Clarifying its boundaries strengthens rather than weakens the framework.

### 9.1. Normative constraints and non relativism

Cultural fidelity does not entail endorsement of all locally dominant norms. Political language technologies operate within broader legal and ethical constraints, including protections for vulnerable groups and prohibitions against incitement. A system that preserves local discourse patterns while amplifying harm would fail democratic safety criteria. Adaptation therefore functions within outer normative boundaries independent of any single political tradition.

### 9.2. Incommensurability and tradeoffs

Political ontologies are often partially overlapping rather than fully commensurable. Concepts such as representation, sovereignty, or legitimacy may carry distinct institutional and historical meanings across contexts. Where conceptual divergence exists, alignment cannot be reduced to translation or parameter tuning; systems should represent uncertainty, surface alternative mappings, or abstain from forced equivalence.

There is also a tradeoff between generality and local calibration. Highly specialized models may achieve stronger ontology level alignment but reduced portability, whereas globally general systems risk flattening culturally salient distinctions. Trustworthy deployment requires explicit documentation of where a system lies on this spectrum and what recalibration is necessary for cross cultural transfer.

### 9.3. Governance capacity and implicit default

Cultural adaptation depends on institutional capacity for documentation, participatory review, monitoring, and remediation. In environments lacking meaningful oversight or appeal mechanisms, even technically sophisticated adaptation strategies may fail to produce trustworthy outcomes.

Non adaptivity, however, is not a neutral baseline. Deploying systems trained primarily on dominant language sources into distinct political contexts implicitly imposes external ontologies and discourse norms. The relevant choice is therefore between explicit, documented adaptation and implicit normative default.

## 10. Conclusion

Culturally adaptive political NLP is a sociotechnical challenge. Large language models can support political analysis, deliberation, and civic technology, but only if they are designed and governed with explicit attention to cultural representation, pragmatic competence, and democratic accountability. Translation alone does not provide cultural adaptation. Trustworthy systems require participatory data and evaluation, selective transfer learning, disaggregated benchmarks, and institutional processes that enable auditing and contestability.

Future work should prioritize culturally grounded benchmarks for political discourse, governance mechanisms that operationalize community influence, and evaluation methods that measure cultural fidelity and democratic safety alongside predictive performance. The central goal is not only better prediction. It is better inclusion, clearer contestability, and fewer representational harms in the political systems where language technologies increasingly operate. Embedding cultural adaptation into benchmark design and shared task infrastructure can make these commitments empirically testable rather than purely aspirational. In doing so, the PoliticalNLP community can align technical innovation with democratic legitimacy across diverse political contexts.

## Limitations

This paper is primarily a conceptual and theoretical contribution. While we provide an illustrative application of the evaluation matrix to cross lingual stance detection and ground our discussion in concrete examples of concept collapse, we do not present a full scale empirical validation of the framework against a deployed system. Such validation would require access to multiple culturally diverse political NLP systems and coordinated evaluation across language communities, which we identify as a near term priority for future work.

The breadth of the framework, spanning technical, governance, and deliberative dimensions, necessarily limits the depth of treatment in each area. The governance and deliberative democracy discussions draw on established literature but do not substitute for the specialized analyses that scholars in those fields would provide. We intend the framework as an integration point rather than a replacement for disciplinary expertise.

Finally, the concrete examples of concept collapse focus on a limited number of political traditions (Senegalese *demokaraasi*, Arabic *shura*,

Swiss multilingual stance). While these illustrate the general mechanisms, additional culturally grounded case studies from other regions and political systems are needed to test the framework's generality and to identify failure modes specific to political traditions not yet examined.

## Ethical Considerations

Political NLP systems operate in domains where errors can affect civic participation, representation, and trust in institutions. Researchers and practitioners working in this space should be attentive to the risks of imposing external political categories on local discourse, the potential for culturally insensitive moderation to silence legitimate political speech, and the asymmetric distribution of both benefits and harms when systems are deployed across unequal power relations.

Participatory dataset development, while valuable, must be implemented with genuine community involvement and meaningful consent, particularly when data involves vulnerable speakers or politically sensitive topics. Tokenistic consultation that uses community involvement as a legitimacy marker without transferring meaningful control does not satisfy the governance requirements outlined in this paper.

We also note that the evaluation matrix itself carries normative commitments, particularly in the democratic safety dimension, which assumes a baseline commitment to inclusive participation and contestability. These commitments may need to be renegotiated when frameworks are applied across political systems with fundamentally different governance assumptions.

## Acknowledgments

This work was made possible by the National Priorities Research Program grant NPRP14C-0916-210015 from the Qatar National Research Fund (QNRF), part of the Qatar Research, Development and Innovation Council (QRDI). The authors also acknowledge the Artificial Intelligence and Media Lab (AIM Lab) at Northwestern University in Qatar (NU-Q) and the MARSAD Lab for providing valuable resources and support that contributed to this research.

## 11. Bibliographical References